\documentclass[letterpaper]{article}
\usepackage[T1]{fontenc}
\usepackage[latin9]{inputenc}
\usepackage[english]{babel}
\usepackage{graphicx}
\usepackage{booktabs}
\usepackage[usenames, dvipsnames]{color}
\usepackage{epsfig}
\usepackage{arxiv}
\usepackage{amssymb}
\usepackage{epstopdf}
\usepackage{algorithm,algorithmic,amsmath}
\usepackage{algorithm,amsmath}
\usepackage{subfig}
\usepackage{amsthm}
\usepackage[title]{appendix}

\usepackage{pdfpages}
\usepackage{ifthen}
\usepackage{authblk}
\usepackage{breakurl}

 \newlength\titlebox 
\makeatletter
\def\BState{\State\hskip-\ALG@thistlm}
\makeatother

\begin{document}
\title{\textbf{Adversarial TCAV - Robust and Effective Interpretation of Intermediate Layers in Neural Networks}}

\author[1\thanks{Untangle AI, 79 Ayer Rajah Crescent, \#03-01, Singapore. Correspondence to: Rahul Soni \textless{}sn.rahul99@gmail.com\textgreater{}}\, ,2]{Rahul Soni}
\author[2]{Naresh Shah}
\author[1]{Chua Tat Seng}
\author[2]{Jimmy D. Moore}
\affil[1]{School of Computing, National University of Singapore}
\affil[2]{Untangle AI, Singapore}
\maketitle

\begin{abstract}
Interpreting neural network decisions and the information learned in intermediate layers is still a challenge due to the opaque internal state and shared non-linear interactions. Although \cite{kim2017interpretability} proposed to interpret intermediate layers by quantifying its ability to distinguish a user-defined concept (from random examples), the questions of robustness (variation against the choice of random examples) and effectiveness (retrieval rate of concept images) remain. We investigate these two properties and propose improvements to make concept activations reliable for practical use.

Effectiveness: If the intermediate layer has effectively learned a user-defined concept, it should be able to recall --- at the testing step --- most of the images containing the proposed concept. For instance, we observed that the recall rate of Tiger shark and Great white shark from the ImageNet dataset with ``Fins'' as a user-defined concept was only $18.35\%$ for VGG16. To increase the effectiveness of concept learning, we propose A-CAV --- the Adversarial Concept Activation Vector --- this results in larger margins between user concepts and (negative) random examples. \textbf{This approach improves the aforesaid recall to $76.83\%$ for VGG16}.

For robustness, we define it as the ability of an intermediate layer to be consistent in its recall rate (the effectiveness) for different random seeds. We observed that \cite{kim2017interpretable} has a large variance in recalling a concept across different random seeds. For example the recall of cat images (from a layer learning the concept of \texttt{tail}) varies from $18\%$ to $86\%$ with $20.85\%$ standard deviation on VGG16. We propose a simple and scalable modification that employs a Gram-Schmidt process to sample random noise from concepts and learn an average ``concept classifier''. \textbf{This approach improves the aforesaid standard deviation from $20.85\%$ to $6.4\%$}.
\end{abstract}

\section{Introduction}
Interpretability in Deep Learning has been gaining traction given that the deep neural networks are being employed in critical applications such as medical diagnostics \cite{zhang2018interpretable, gale2018producing} or autonomous vehicles \cite{kim2017interpretable, kim2017end} among numerous other domains. Given the wide applications of neural networks, it is important to have model explanation systems that provide a means to debug the model and establish trust for production usage. For example, an explanation technique that tells which pixels are maximally activated for a particular prediction is one way to assess that the model is right for right reasons.

To provide explanations that are useful in probing and improving the model further, it is important that, (1) the explanation technique allows understanding of an object by interpreting its parts --- similar to the way humans understand it -- for example, a scuba diver wears different parts such as ``oxygen regulator'', ``snorkel mask'', ``pressure guage'' etc., and (2) the explanation technique outlines \textit{where} in the model the concept was being learned --- for example, in a face detection task \cite{farfade2015multi}, knowing that the $k^{th}$-layer of a CNN learns the concept of color will help in removing the color bias from the network.

To tackle the aforesaid challenges, Kim \cite{kim2017interpretable} proposes one such approach --- Concept Activation Vector (CAV) --- that provides layer-level understanding of a user-defined concept. CAV requires a small set of examples images of concepts that are easily understood to humans, and, for each layer in network, learns an activation vector representative of that concept. Specifically, given a small set of ``concept examples'', CAV generates random vectors, called ``non-concept examples''; passes the two sets of examples through the network; collects layer activations, and builds a binary linear classifier. The coefficients of the linear model are then defined to be the representative of such a concept.

Although TCAV is a great approach to probe into intermediate layers of a neural network, its accuracy is low in practice, and requires many iterations to have right samples of ``random examples'' that represent most of the things about ``non-concepts''. We carry forward Kim's \cite{kim2017interpretable} work on testing with CAV to make it more effective (increase empirical performance) and robust to external variations such as the choice of random sampling of non-concept examples. 

We pursue our study of CAVs to address the following objectives:
\begin{itemize}
    \item \textbf{Effectiveness}: The effectiveness or strength of the CAV in learning a concept is determined by its strength in retrieving test set images (with known ground truth) containing the said concepts. We propose to improve the retrieval rate by separating the positive and negative examples farther away from the linear decision boundary. We call this method ``Adversarial CAV'' (A-CAV) and testing with this method --- A-TCAV. Next, we transform the negative examples to learn a disjoint subspace of positive and negative data to prevent non-linearity. We call this method ``Orthogonal Adversarial CAV'' (OA-CAV) and testing with this method --- ``OA-TCAV''.
    \item \textbf{Robustness}: TCAV performance significantly varies with changing the random seed since the random samples can take arbitrary shape including, but not limited to, overlapping with the concept activations. We propose to improve the robustness in CAV by learning coefficient vectors from multiple linear models and computing CAV in the direction orthogonal to the centroid of those coefficient vectors.
    \item \textbf{Prevention against adversarial attacks}: We empirically demonstrate the powerful side-effect of the proposed A-CAV in preventing adversarial attacks at \textit{intermediate layers}. Since the A-CAV is learned from positive adversarial perturbations (that move data points towards the higher Softmax score), we observe that this step suppresses the effect of follow-up adversarial attacks at the testing phase.
    \item \textbf{Investigating Bias in the intermediate layers}: Through our experimental results, we analyze if the model's bottleneck layers are biased towards texture over shape or vice versa.
\end{itemize}

In the rest of the paper, we empirically demonstrate the effect of simple modifications proposed in this work by defining several human level concepts, and performing experiments with sequential and non-sequential network architectures.

\section{Related work}
Previous methods have shown interpretability in neural networks, either by (1) updating models to yield interpretability \cite{ustun2013supersparse, caruana2015intelligible} or (2) explaining models in-lieu of simpler / self-explaining examples (\cite{ross2017right, lundberg2017unified, koh2017understanding}), or (3) generating explanations from a trained neural network (\cite{springenberg2014striving,zeiler2014visualizing,kim2016examples,lundberg2017unified,koh2017understanding, springenberg2014striving, selvaraju2017grad, sundararajan2017axiomatic, smilkov2017smoothgrad, kindermans2017learning}). Given that models have become complex, large, and take a long time to train, the later approach to interpretability has become prominent in recent years.

In category (1), \cite{ross2017right}, for instance, attempts to explain model predictions by generating alternate, ``explainable'' input vectors which are binary masks that depict either a presence or absence of a feature. \cite{koh2017understanding} attempts to identify which training examples maximally influence the prediction of a given test point. One of the positive side effects of Koh's work \cite{koh2017understanding} was to use influence functions to sort the most relevant training examples, or, remove bad training examples from the training set as a post augmentation process.

To generate explanations from a trained neural network, a typical approach is to observe the effect of model on its prediction when the input pixels are perturbed \cite{springenberg2014striving}. One of the enhancements to such perturbations was Selvaraju's work \cite{selvaraju2017grad} --- GradCAM --- where the last convolution layer is perturbed instead of the input layer followed by linearly pooling the gradients from all channels. This gradient is then reshaped to the input dimension to generate explanations. \cite{sundararajan2017axiomatic, smilkov2017smoothgrad} proposed a meta attribution algorithm that can sit on top of existing perturbation based algorithms and helps to suppress noise. \cite{kindermans2017learning} proposed a method to learn the noise (called distraction) and filter it out at the time of generating explanations. \cite{kindermans2017learning} is the most recent work on suppressing distractions in the input signal.

\section{Adversarial TCAV}
\begin{figure}[!htb]
\vskip 0.2in
\begin{center}
\centerline{\includegraphics[width=0.75\linewidth, height=5cm]{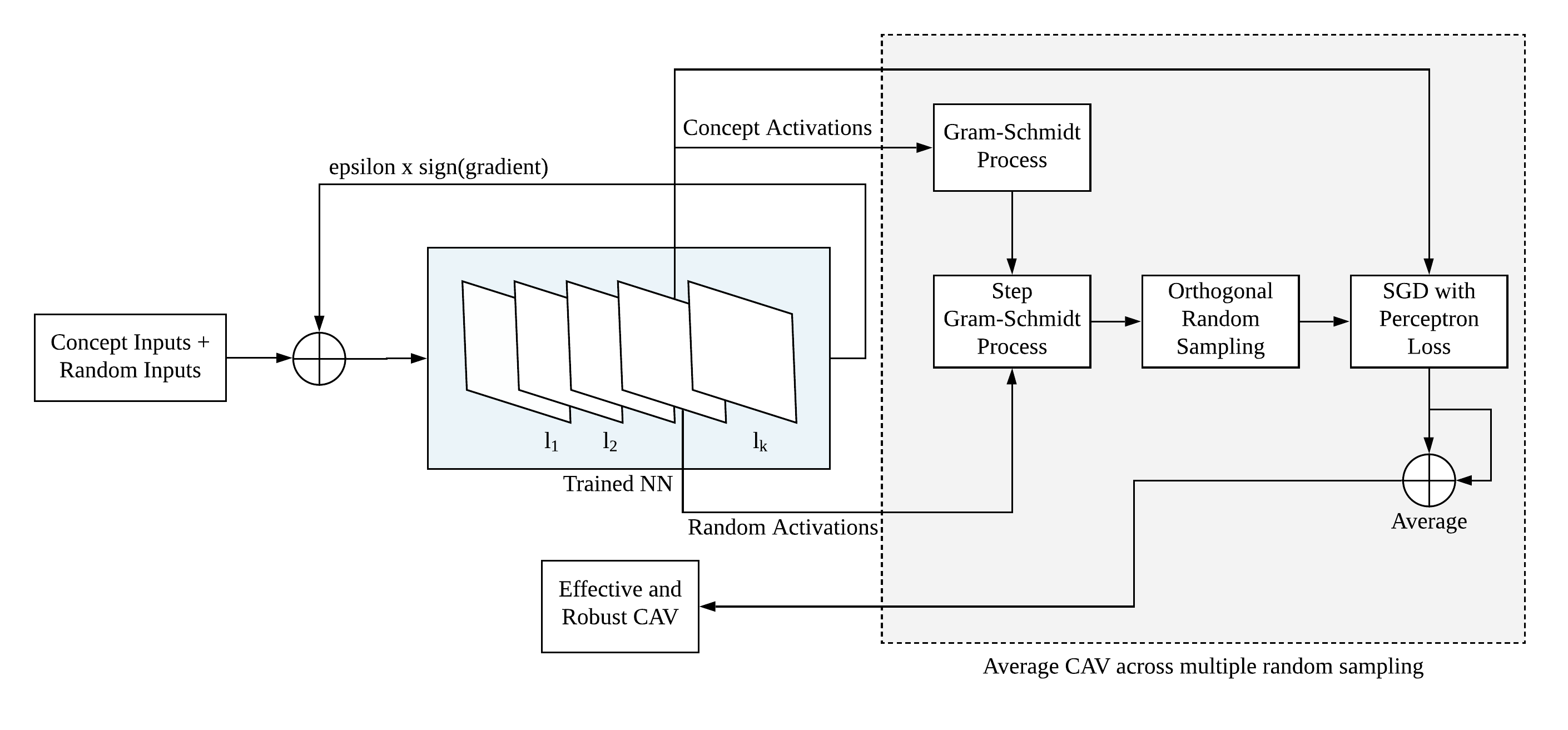}}
\caption{Proposed improvements to CAV. Both, concept inputs and random inputs are updated with adversarial perturbation first, before collecting their activations at a given layer. A repeated Gram-Schmidt process and linear modelling re-computes the random activations and CAV, respectively, to obtained adversarial and orthogonal CAVs}
\label{fig:architecture}
\end{center}
\vskip -0.2in
\end{figure}

In this section, we investigate the mechanics of Concept Activation Vectors (CAV) \cite{kim2017interpretability} and propose improvements to make it more effective and robust. The overall architecture is outlined in Fig~\ref{fig:architecture}.

Standard, gradient based, explanation techniques \cite{springenberg2014striving, selvaraju2017grad, chattopadhay2018grad, sundararajan2017axiomatic, zeiler2014visualizing} measure the sensitivity of the model prediction w.r.t the input tensor. Kim's work \cite{kim2017interpretability} proposes conceptual sensitivity - the directional input sensitivity projected along the direction of the Concept Activation Vector (CAV).

To compute such a directional derivative, the CAV is computed first, as follows: Given a small set of example images of a concept, $C$ (for instance, ``fins'' as a concept to retrieve shark images), a binary dataset is constructed containing activations of concepts as positive examples and activations of non-concepts (random vectors) as negative examples. A concept activation vector, ${\boldsymbol{v}^l}_C$ at layer $l$ is then defined as the normal vector to the linear decision boundary separating the binary dataset. 

For an input test point $\boldsymbol{x} \in \mathbb{R}^n$, let $f_l(\boldsymbol{x}) \in \mathbb{R}^m$ be the activation at layer $l$ and let $h_{l,k}$ be the mapping ($\mathbb{R}^m \rightarrow \mathbb{R}$) of activation $f_l(\boldsymbol{x})$ to logit value of $k^{th}$ prediction class. The conceptual sensitivity, $S_{C,k,l}$ w.r.t concept $C$ is then defined as the projection of the gradient of input at that layer onto the concept activation vector\footnote{As noted in \cite{kim2017interpretability}, the Conv layer activation tensors are $4$-dimensional ($H,W,C_{in}, C_{out}$). In this case, we flatten the output to obtain an activation vector of size ($H * W * C_{in} * C_{out}$)}:
\begin{equation}\label{eq:1}
    \begin{split}
     S_{C,k,l} &= \nabla_{h_{l,k}} (f_l(\boldsymbol{x})) \cdot {\boldsymbol{v}^l}_C
    \end{split}
\end{equation}

To make the CAVs of great practical use, for instance to retrieve all images containing a given concept from a large unlabelled corpora, we need to assess the effectiveness (higher retrieval rate) and robustness (consistent retrieval rate) of the method.

\subsection{Effectiveness in TCAVs} From conceptual sensitivity \ref{eq:1}, it is evident that the effectiveness of TCAV is dependent on the ability of a linear classifier to separate concepts from random activations. As the choice of sampling random tensors is not restricted to a particular subpspace, (for example, sampling from outside the subspace of concept activations), the random activations can make the resulting dataset non-linear, thus resulting in poorer retrieval rate (as observed empirically).

\subsubsection{Adversarial Separation} To prevent this uncontrolled overlap of activation vectors, we propose to push the activation vectors in their respective category, away from the decision boundary.
This can be achieved by adversarial separation - adding a small perturbation in the input vector along the direction of its gradient.

\begin{equation}\label{eq:2}
    \begin{split}
     \boldsymbol{x}^{new} &= \boldsymbol{x} + \epsilon \textit{sign} \left( \nabla_{h_{0,k}} (f_0(\boldsymbol{x})) \right)
    \end{split}
\end{equation}

where $h_{0,k}$ is the mapping ($\mathbb{R}^n \rightarrow \mathbb{R}$) from input $\boldsymbol{x}$ to output prediction $k$.

Adding such a perturbation increases the Softmax score for the predicted class. Since the positive and negative activations predict different classes (naturally), this increases the separation between the two activations resulting in large margin decision boundary. Such adversarial approach is not new and has been shown to have a large positive effect in separating data points
\cite{goodfellow2014explaining, liang2017enhancing, lee2018simple}.

\subsubsection{Disjoint Concept Subspaces}
Lee and Seung \cite{lee1999learning} in their \texttt{Nature} article propose Non-Negative Matrix factorization (NMF) (and, later, \cite{donoho2004does} proposes geometrical interpretations) as a way to learn parts of an object (called the basis vectors). 

\begin{figure}[!htb]
\vskip 0.2in
\begin{center}
\centerline{\includegraphics[width=0.75\linewidth, height=8cm]{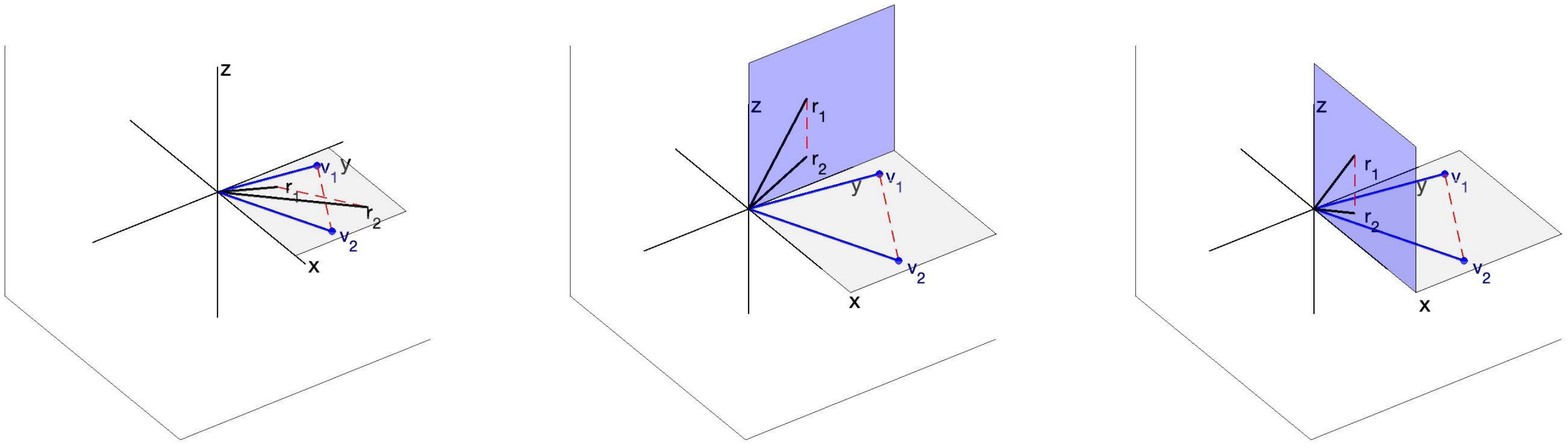}}
\caption{Examples of random sampling. \textbf{left}: no constraints on random sampling can lead to overlapping points or distributions with smaller decision margin, \textbf{middle, right}: two instances of the proposed random sampling using the Gram-Schmidt process. $\textbf{r}_1$, $\textbf{r}_2$ represent random activations and $\textbf{v}_1$, $\textbf{v}_2$ concept activations in a $2D$ subspace.}
\label{fig:projection}
\end{center}
\vskip -0.2in
\end{figure}

We employ a procedure similar to NFM to further enhance the separability of concept activations from non-concept activations. Specifically, we run a two-step Gram-Schmidt Orthogonalization process (\cite{bjorck1994numerics}), where, in the first step, we compute an orthogonal basis of the subspace spanned by concept activation vectors --- called the \texttt{concept basis}. In the second step, we compute another orthogonal basis --- called the \texttt{non-concept orthogonal basis} (or, without loss of generality, \texttt{non-concept basis}) - which is disjoint to the \texttt{concept basis}. With concept activations as positive examples, we can now generate negative examples for the downstream binary classifier by sampling activations from the \texttt{non-concept basis}. Our empirical observations suggests that sampling from more than one instances of \texttt{non-concept basis} provides good generalization. An example of such a sampling is shown in Fig~\ref{fig:projection}. Algorithm~\ref{alg:adv} and Algorithm~\ref{alg:advOrth} demonstrate the two modifications to CAV respectively.

\begin{algorithm}[!htb]
   \caption{Adversarial Separation} 
   \label{alg:adv}
\begin{algorithmic}
   \STATE {\bfseries Input:} Concept examples $\boldsymbol{x}_i \in \mathbb{R}^n$, trained Neural Network estimator, $f(\cdot)$
   \STATE Initialize $\boldsymbol{r}_i \in \mathbb{R}^n \longleftarrow \texttt{random tensors}$
   \STATE Initialize $K \longleftarrow \texttt{\# concept examples}$
   \STATE Initialize $L \longleftarrow \texttt{\# non-concept examples}$
   \STATE Initialize $l \longleftarrow \texttt{neural network layer index}$
   \STATE Initialize $\texttt{concept activations, } \texttt{cact} \longleftarrow \texttt{\{\}}$
   \STATE Initialize $\texttt{non-concept activations, }$
   \STATE $\texttt{ncact} \longleftarrow \texttt{\{\}}$
   \STATE Initialize $g(\cdot) \longleftarrow \texttt{SGD Classifier with}$
   \STATE $\texttt{perceptron loss}$
   \FOR{$i=1$ {\bfseries to} $K$}
   \STATE $\boldsymbol{x}_i \longleftarrow \boldsymbol{x}_i + \epsilon \textit{sign} \left( \nabla_{h_{0,k}} (f_0(\boldsymbol{x}_i)) \right)$
   \STATE $\texttt{cact}.insert(f_l(\boldsymbol{x}_i))$
   \ENDFOR
   
   \FOR{$i=1$ {\bfseries to} $L$}
   \STATE $\boldsymbol{r}_i \longleftarrow \boldsymbol{r}_i + \epsilon \textit{sign} \left( \nabla_{h_{0,k}} (f_0(\boldsymbol{r}_i)) \right)$
   \STATE $\texttt{ncact}.insert(f_l(\boldsymbol{r}_i))$
   \ENDFOR
   \STATE $\texttt{TRAIN} \,\,\, g(\texttt{cact}, \texttt{ncact})$
   \STATE ${\boldsymbol{u}^l}_C \longleftarrow \texttt{coefficients of g(cact, ncact)}$
\STATE ${\boldsymbol{v}^l}_C \longleftarrow -1 * {\boldsymbol{u}^l}_C$
\end{algorithmic}
\end{algorithm}

\begin{algorithm}[!htb]
   \caption{Orthogonal Adversarial Separation. Here, \texttt{GRAM-SCHMIDT} denotes the Gram-Schmidt Orthogonalization (GSO) process and \texttt{STEP GRAM-SCHMIDT} is a single step of GSO that returns a component of input vector orthogonal to the given basis set.}
   \label{alg:advOrth}
\begin{algorithmic}
   \STATE {\bfseries Input:} Concept examples $\boldsymbol{x}_i \in \mathbb{R}^n$, trained Neural Network estimator, $f(\cdot)$
   \STATE Initialize $\boldsymbol{r}_i \in \mathbb{R}^n \longleftarrow \texttt{random tensors}$
   \STATE Initialize $K \longleftarrow \texttt{\# concept examples}$
   \STATE Initialize $L \longleftarrow \texttt{\# non-concept examples}$
   \STATE Initialize $l \longleftarrow \texttt{neural network layer index}$
   \STATE Initialize $\texttt{concept activations, } \texttt{cact} \longleftarrow \texttt{\{\}}$
   \STATE Initialize $\texttt{non-concept activations, }$
   \STATE $\texttt{ncact} \longleftarrow \texttt{\{\}}$
   \STATE Initialize $g(\cdot) \longleftarrow \texttt{SGD Classifier with}$
   \STATE $\texttt{perceptron loss}$
   \FOR{$i=1$ {\bfseries to} $K$}
        \STATE $\boldsymbol{x}_i \longleftarrow \boldsymbol{x}_i + \epsilon \textit{sign} \left( \nabla_{h_{0,k}} (f_0(\boldsymbol{x}_i)) \right)$
        \STATE $\texttt{cact}.insert(f_l(\boldsymbol{x}_i))$
   \ENDFOR
   \FOR{$i=1$ {\bfseries to} $L$}
        \STATE $\boldsymbol{r}_i \longleftarrow \boldsymbol{r}_i + \epsilon \textit{sign} \left( \nabla_{h_{0,k}} (f_0(\boldsymbol{r}_i)) \right)$
        \STATE $\texttt{ncact}.insert(f_l(\boldsymbol{r}_i))$
   \ENDFOR
   
   \STATE $\texttt{concept\_basis} \longleftarrow \texttt{GRAM-SCHMIDT(cact)}$
   \FOR{$i=1$ {\bfseries to} $L$}
        \STATE $\boldsymbol{q}_i \longleftarrow \texttt{ncact.get(i)}$
        \STATE ${\boldsymbol{q}_i}^\perp \longleftarrow \texttt{STEP GRAM-SCHMIDT(cact, }\boldsymbol{q}_i)$
   \ENDFOR
   \STATE $\texttt{non\_concept\_basis} \longleftarrow \texttt{GRAM-SCHMIDT(\{${\boldsymbol{q}_1}^\perp, {\boldsymbol{q}_2}^\perp \cdots {\boldsymbol{q}_L}^\perp $\})}$
   \STATE $\texttt{ncact} \longleftarrow \texttt{sample(non\_concept\_basis)}$
   \STATE $\texttt{TRAIN} \,\,\, g(\texttt{cact}, \texttt{ncact})$
   \STATE ${\boldsymbol{u}^l}_C \longleftarrow \texttt{coefficients of g(cact, ncact)}$
\STATE ${\boldsymbol{v}^l}_C \longleftarrow -1 * {\boldsymbol{u}^l}_C$
\end{algorithmic}
\end{algorithm}

\subsection{Robustness in TCAVs}
We performed several validations of CAV (across multiple random seeds and multiple concepts) to sort images in relation to a given concept. We observed that TCAV was not effective consistently, i.e., it had a high standard deviation in recall rate. For instance, to sort Tiger shark and Great white shark images from the Imagenet validation categories in relation to \texttt{fins} concept, the recall rate varied from $19\%$ to $78\%$ with standard deviation of $19.03\%$.

To make CAVs robust to random seed selection, we propose to run multiple instances of linear model on different random examples sampled from the \texttt{non-concept basis}. CAV, then points in the centroid direction of the learned linear coefficient vectors. The number of such draws is a fixed, configurable hyper-parameter, $N_d$. This modification is surprisingly effective - especially when combined with the adversarial and orthogonal separation. We observe that the approach is not only robust against changing seeds, it is also robust against changing the hyper-parameter, $N_d$ itself. For example, changing the hyper-parameter from $N_d = 10$ to $N_d = 100$ does not yield additional performance gains (reduction in standard-deviation) --- thus promising a stable result consistently.

\section{Experimental Setup}
Without loss of generality, we validate our method on a subset of 20 Imagenet categories --- which we call the ``Imagenet20 Dataset'' --- to simplify the CAV generation process. We test our method on modified ``VGG16'', ``Resnet18'', and ``Alexnet'' --- where the last layer is modified to output 20 class probabilities.

\subsection{Models}
We fine-tune the modified models to Imagenet20 dataset using the $80-20$ train-validation split, with the Adam optimizer, for $50$ epochs, and learning rate decay when no validation loss improvement over a patience of $10$ epochs. The initial learning rate is set to $0.001$ and reduced by a factor of $0.25$ after $patience=10 \,\, \textit{epochs}$ is reached. Table~\ref{tab:finetune} summarizes the training statistics of the three models.

\begin{table}[!htb]
\caption{Training summary of modified VGG16 (\textbf{VGG16*}), modified Resnet18 (\textbf{Resnet18*}), and modified Alexnet (\textbf{Alexnet*}) after the fine-tuning step. Each model is first loaded with pretrained Imagenet weights and then finetuned (for 50 epochs) to output 20 class probabilities.}
\label{tab:finetune}
\vskip 0.15in
\begin{center}
\begin{small}
\begin{sc}
\resizebox{0.47\textwidth}{!}{
\begin{tabular}{ccccccc}
\hline
 & \multicolumn{2}{c}{\textbf{VGG16*}} & \multicolumn{2}{c}{\textbf{Resnet18*}} & \multicolumn{2}{c}{\textbf{Alexnet*}} \\ \hline
 & Loss & \multicolumn{1}{c|}{Acc} & Loss & \multicolumn{1}{c|}{Acc} & Loss & Acc \\
\textbf{Train} & 0.08 & \multicolumn{1}{c|}{96.95} & 0.001 & \multicolumn{1}{c|}{99.98} & 0.098 & 97.13 \\
\textbf{Valid} & 0.121 & \multicolumn{1}{c|}{95.68} & 0.56 & \multicolumn{1}{c|}{85.55} & 0.33 & 88.65 \\
\textbf{Test} & 0.28 & \multicolumn{1}{c|}{91.2} & 0.74 & \multicolumn{1}{c|}{80.70} & 0.456 & 84.70\\
\bottomrule
\end{tabular}}
\end{sc}
\end{small}
\end{center}
\vskip -0.1in
\end{table}

\subsection{Dataset}
To construct Imagenet20, we selected categories such that more than one category shares a common user defined concept, for example: the \texttt{fins} concept is common to \texttt{Great white shark} and \texttt{Tiger shark}. The concept of  \texttt{tail} was common to \texttt{Egyptian cat and Tiger cat}. This was done to ensure that CAV learns a concept and does not simply do pattern matching with the context. For example, the \texttt{Tiger cat} validation dataset of Imagenet contains $24\%$ actual tiger images ($12$ out of $50$ images). Naturally, the context (background) of these images depicts ``outdoor'' setting, which is in contrast to the \texttt{Egyptian cat} category where the context is mostly indoors. Following list of concepts are constructed and validated for our method:
\begin{itemize}
    \item \textbf{Fins} - To recall images of \texttt{Great white shark} and \texttt{Tiger shark} from the Imagenet20 validation set.
    \item \textbf{Cat's fur} - To recall images of \texttt{Egyptian cat} and \texttt{Tiger cat} from the Imagenet20 validation set.
    \item \textbf{Cat's tail} - To recall images of \texttt{Egyptian cat} and \texttt{Tiger cat} from the Imagenet20 validation set.
\end{itemize}

\subsection{Parameter Setting}
To compute CAV, we used the SGD linear model with perceptron loss and constant learning rate (to avoid overfitting the small concept dataset). We computed both the baseline CAV and our method (A-CAV, OA-CAV) for $50$ random seeds and compared recall results. For the adversarial separation in our method, the $\epsilon$ term is set to a tune-able hyper-parameter taking values in $\{0.1, 0.01, 0.005, 0.001, 0.0001\}$. We computed $3$ instances of the Grahm-Schmidt Orthogonalization (GSO) process to obtain the ``non-concept activations''. The linear model takes the concept activations and the output of GSO to learn a CAV. We performed $10$ such iterations of linear model to obtain an ``Adversarial CAV'' (A-CAV).

\section{Results and Insights}
We summarize our experimental results in this section and shed light on the following insights:
\begin{itemize}
    \item Is the bottleneck Convolutional layer biased towards shape or textures?
    \item Effect of adversarial attacks on shapes versus textures
\end{itemize}

\subsection{Image Retrieval (Recall rate)}
We use the learned A-CAV and OA-CAv and test its strength by sorting images of the Imagenet20 test dataset. For each image, we compute the test activation vector at the same layer on which the CAV was built and compute the inner product according to Eq~\ref{eq:1}. We summarize the recall results of different models tested for different concepts in Table~\ref{tab:vgg},~\ref{tab:resnet},~\ref{tab:alex}. In the tables, we refer to \textbf{TCAV} as the baseline method \cite{kim2017interpretable} test, \textbf{A-TCAV} --- the proposed adversarial CAV test, and \textbf{OA-TCAV} --- the proposed A-TCAV test with orthogonal random sampling of ``non-concept'' activations. We observed that the proposed methods are consistent across $50$ random seeds with significantly lower standard deviation in recall compared to baseline TCAV. 

We observe that A-CAV performs consistently well across all formats with occasional higher performance from OA-TCAV (testing \texttt{fins} on VGG16, \texttt{fur} on Alexnet). This is due to the reason that OA-CAV learns orthogonal disjoint subspaces which relaxes the degree of freedom of the linear decision boundary. Nevertheless, we believe that OA-CAV based sampling provides meaningful insights and opens possibilities for future investigations.

\begin{figure}[t!]
    \centering
    \subfloat[\textbf{VGG16}]{{\includegraphics[width=0.47\linewidth, height=7cm]{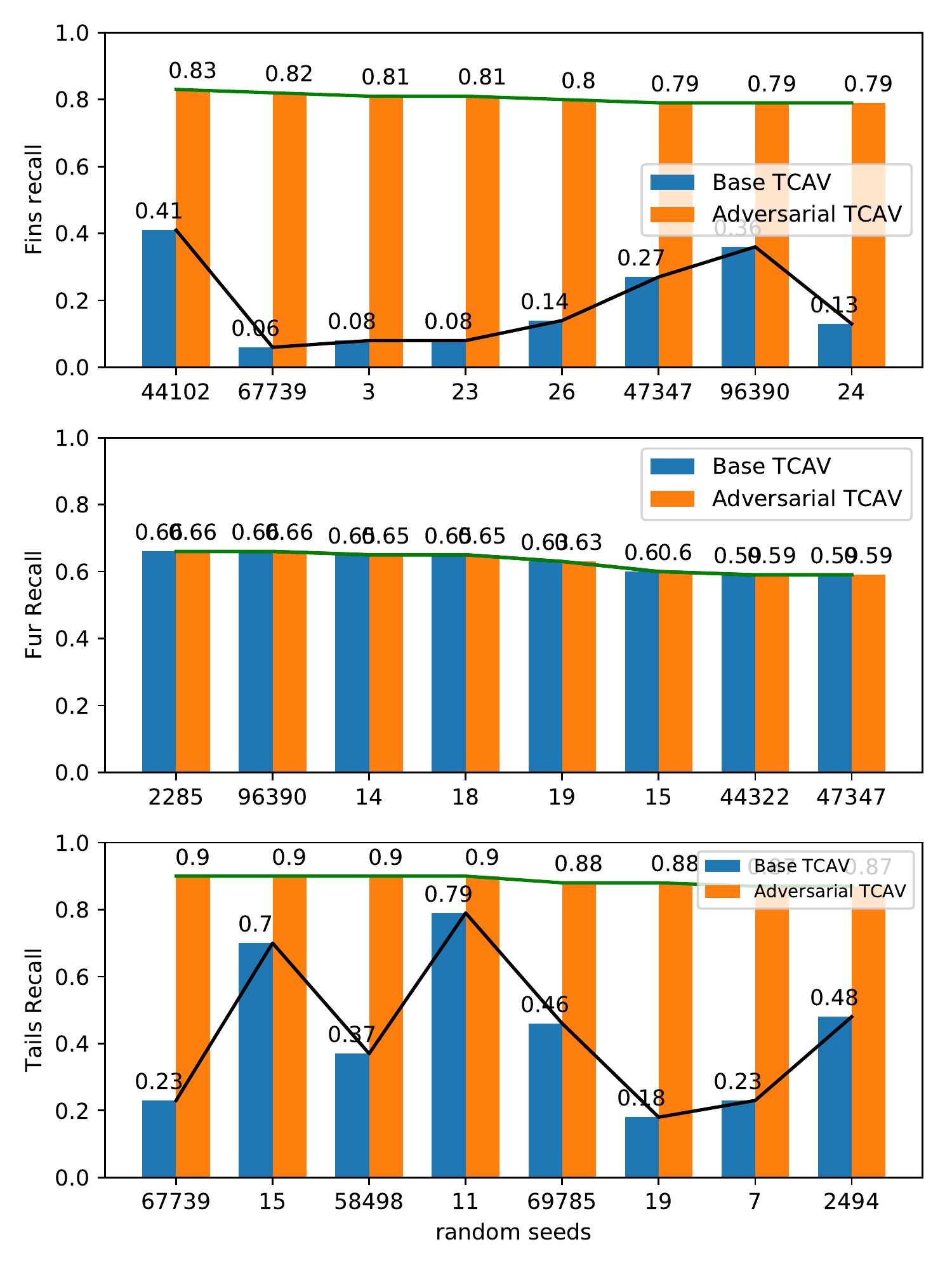} }}%
    \qquad
    \subfloat[\textbf{Resnet18}]{{\includegraphics[width=0.47\linewidth, height=7cm]{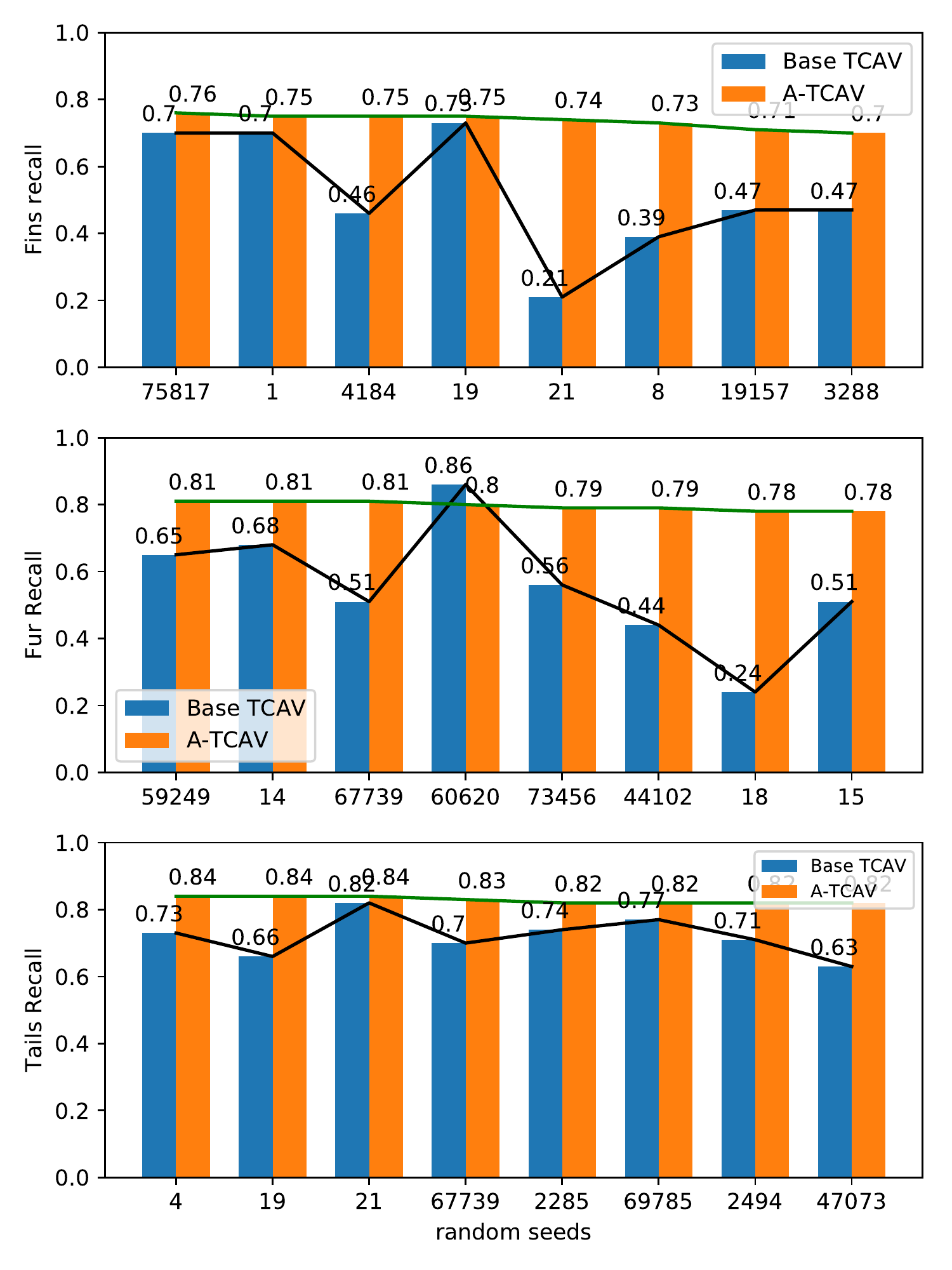} }}\\
    \qquad    
    \subfloat[\textbf{Alexnet}]{{\includegraphics[width=0.47\linewidth, height=7cm]{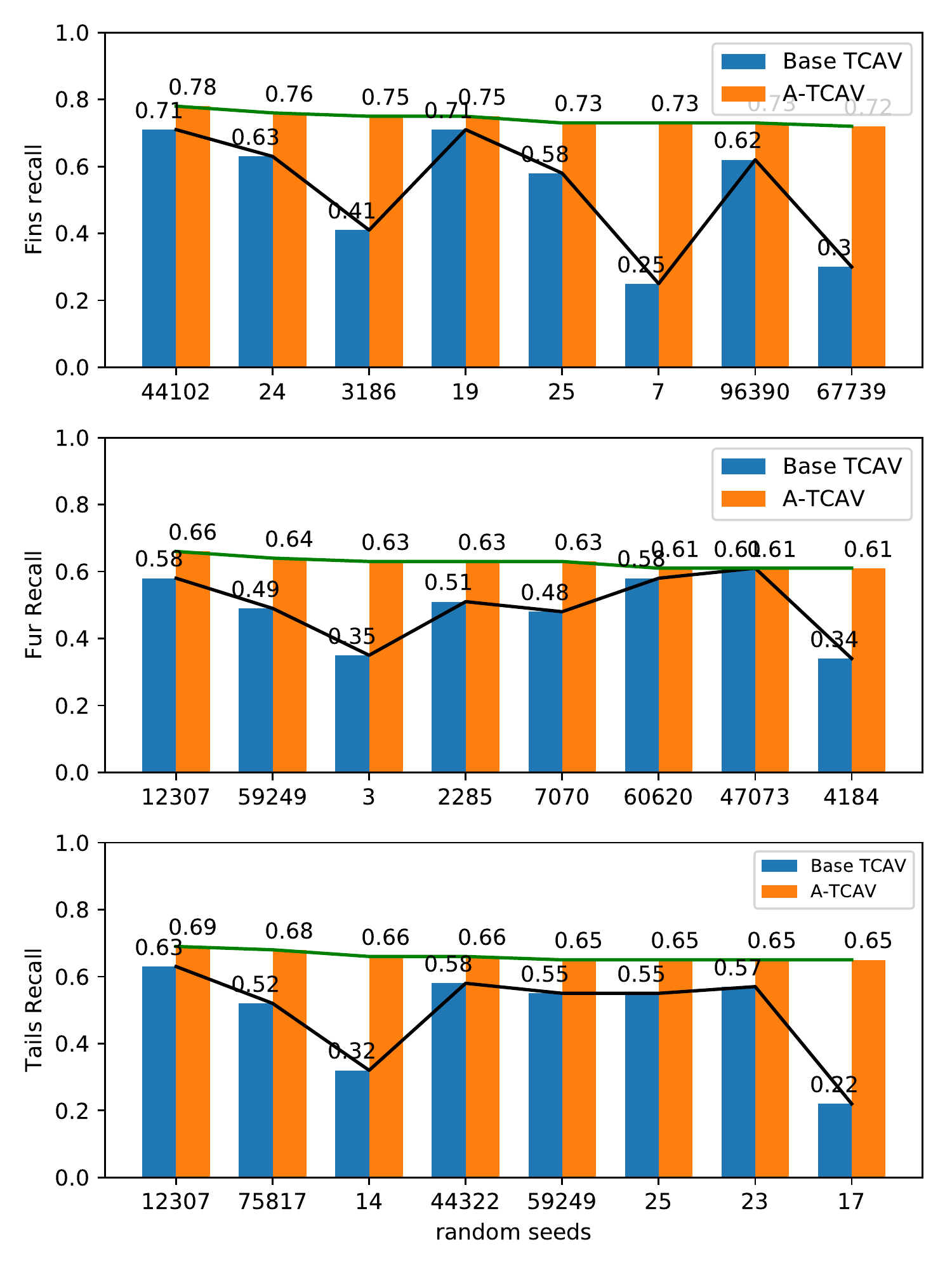} }}%
\caption{Comparison of recall rate of TCAV, A-TCAV, and OA-TCAV; and variation across different seeds (seeds that have resulted in higher accuracy for A-CAV). We observe that the A-TCAV is both effective (higher recall) and robust (consistent to random sampling) compared to the baseline}
\label{fig:barplot}
\end{figure}

Fig~\ref{fig:barplot} shows a trend in the recall rate for a sample of seeds that have shown better performance for A-CAV. The occasional jumps in recall rate of baseline CAV is attributed to a \textit{good} sampling of random vectors that inherently tilts the dataset towards linear separability.

\begin{table*}[!htb]
\caption{\textbf{VGG16 results}. Recall rate of testset images from the Imagenet20 dataset containing a given concept when CAVs were trained on the bottleneck layer Numbers marked in bold represent the best performance and CAVs were learned on the bottleneck layer}
\label{tab:vgg}
\vskip 0.15in
\begin{center}
\begin{small}
\begin{sc}
\begin{tabular}{cccccccccc}
\hline
 & \multicolumn{3}{c}{\textbf{Fins Recall (in \%)}} & \multicolumn{3}{c}{\textbf{Tail Recall (in \%)}} & \multicolumn{3}{c}{\textbf{Fur Recall (in \%)}} \\ \hline
\textbf{} & TCAV & A-TCAV & \multicolumn{1}{c|}{OA-TCAV} & TCAV & A-TCAV & \multicolumn{1}{c|}{OA-TCAV} & TCAV & A-TCAV & OA-TCAV \\
mean & 18.35 & 76.83 & \multicolumn{1}{c|}{\textbf{79.66}} & 51.92 & \textbf{80.60} & \multicolumn{1}{c|}{67.08} & 42.92 & \textbf{56.36} & 50.96 \\
std & 11.08 & \textbf{3.44} & \multicolumn{1}{c|}{3.49} & 20.85 & \textbf{6.4} & \multicolumn{1}{c|}{12.88} & 10.55 & \textbf{3.98} & 7.42 \\
min & 1 & 66 & \multicolumn{1}{c|}{70} & 18.0 & 62.0 & \multicolumn{1}{c|}{27.0} & 20.0 & 50.0 & 32.0 \\
max & 41 & 83 & \multicolumn{1}{c|}{85} & 86.0 & 90.0 & \multicolumn{1}{c|}{88.0} & 62.0 & 66.0 & 63.0\\
\bottomrule
\end{tabular}
\end{sc}
\end{small}
\end{center}
\vskip -0.1in
\end{table*}

\begin{table*}[!htb]
\caption{\textbf{Resnet18 results}. Recall rate of testset images containing a given concept from the Imagenet20 dataset when CAVs were trained on the bottleneck layer. Numbers marked in bold represent the best performance.}
\label{tab:resnet}
\vskip 0.15in
\begin{center}
\begin{small}
\begin{sc}
\begin{tabular}{cccccccccc}
\hline
 & \multicolumn{3}{c}{\textbf{Fins Recall (in \%)}} & \multicolumn{3}{c}{\textbf{Tail Recall (in \%)}} & \multicolumn{3}{c}{\textbf{Fur Recall (in \%)}} \\ \hline
\textbf{} & TCAV & A-TCAV & \multicolumn{1}{c|}{OA-TCAV} & TCAV & A-TCAV & \multicolumn{1}{c|}{OA-TCAV} & TCAV & A-TCAV & OA-TCAV \\
mean & 43.89 & 62.34 & \multicolumn{1}{c|}{\textbf{66.51}} & 67.26 & \textbf{77.11} & \multicolumn{1}{c|}{72.87} & 56.11 & 71.83 & \textbf{76.92} \\
std & 19.28 & \textbf{6.04} & \multicolumn{1}{c|}{12.29} & 14.79 & \textbf{4.17} & \multicolumn{1}{c|}{6.67} & 17.87 & \textbf{4.62} & 8.11 \\
min & 3.0 & 38.0 & \multicolumn{1}{c|}{35.0} & 3.0 & 69.0 & \multicolumn{1}{c|}{41.0} & 12.0 & 62.0 & 49.0 \\
max & 77.0 & 76.0 & \multicolumn{1}{c|}{84.0} & 85.0 & 84.0 & \multicolumn{1}{c|}{84.0} & 86.0 & 81.0 & 89.0\\
\bottomrule
\end{tabular}
\end{sc}
\end{small}
\end{center}
\vskip -0.1in
\end{table*}

\begin{table*}[!htb]
\caption{\textbf{Alexnet results}. Recall rate of testset images containing a given concept from the Imagenet20 dataset when CAVs were trained on the bottleneck layer. Numbers marked in bold represent the best performance.}
\label{tab:alex}
\vskip 0.15in
\begin{center}
\begin{small}
\begin{sc}
\begin{tabular}{cccccccccc}
\hline
 & \multicolumn{3}{c}{\textbf{Fins Recall (in \%)}} & \multicolumn{3}{c}{\textbf{Tail Recall (in \%)}} & \multicolumn{3}{c}{\textbf{Fur Recall (in \%)}} \\ \hline
\textbf{} & TCAV & A-TCAV & \multicolumn{1}{c|}{OA-TCAV} & TCAV & A-TCAV & \multicolumn{1}{c|}{OA-TCAV} & TCAV & A-TCAV & OA-TCAV \\
mean & 36.87 & 64.72 & \multicolumn{1}{c|}{\textbf{65.56}} & 41.79 & \textbf{60.06} & \multicolumn{1}{c|}{47.31} & 42.92 & 56.36 & \textbf{50.96} \\
std & 22.16 & 11.02 & \multicolumn{1}{c|}{\textbf{7.58}} & 14.56 & \textbf{5.23} & \multicolumn{1}{c|}{11.09} & 10.55 & \textbf{3.98} & 7.42 \\
min & 6.0 & 24.0 & \multicolumn{1}{c|}{35.0} & 17.0 & 47.0 & \multicolumn{1}{c|}{24.0} & 20.0 & 50.0 & 32.0 \\
max & 71.0 & 78.0 & \multicolumn{1}{c|}{75.0} & 63.0 & 69.0 & \multicolumn{1}{c|}{70.0} & 62.0 & 66.0 & 63.0\\
\bottomrule
\end{tabular}
\end{sc}
\end{small}
\end{center}
\vskip -0.1in
\end{table*}

\subsection{Bias in the Bottleneck Conv Layer}
\cite{geirhos2018imagenet} argues that neural networks are more reliant on texture than shape to recognize an object. Without having access to the internal working mechanisms of a model, it is hard to comment if the model is globally more reliant on texture or only its localized neuron groups. To resolve this issue, our data construction and experiments provide great insights.

We note that the model's testset recall (in relation to a \textit(conceptual-layer)) is achieved by sorting the sensitivity scores which are achieved by inner product of CAV and a given test example. This linear operation makes the recall rate proportional to the concept sensitivity.

From Table~\ref{tab:vgg},~\ref{tab:resnet},~\ref{tab:alex} which summarizes retrieval results from the \textit{bottleneck layer} of VGG16, Resnet18, and Alexnet, we observe that the recall rate of ``Egyptian cat'' and ``Tiger cat'' in relation to the concept \texttt{fur} is lower than recall rate in relation to the concept of \texttt{tail}. 

We therefore hypothesize that: \textit{the bottleneck layers are less sensitive to learning texture based concepts} (in this case \texttt{cat's fur}) \textit{and more sensitive to shape based concepts} (in this case \texttt{fins}, \texttt{tail}).

\subsection{Prevention Against Adversarial Attacks}
Adversarial attacks have grown to be prominent in deceiving models prediction\cite{goodfellow2014explaining}. As a natural extension, we pose the following question. \textit{What is the effect of adversarial attacks on the concept understanding of intermediate layers?}

Consider a concept $C$ learned at layer $l$ with a given sensitivity score, $S_{C,l}$ for a set of test images. If we replace the testing set with its adversarial equivalent, the distance of testing images from the concept activation would increase and would continue to increase with increasing levels of adversarial perturbation. A method immune to adversarial attack must separate slowly from the CAVs.

\begin{figure}[t!]
    \centering
    \subfloat[seed=14]{{\includegraphics[width=0.46\linewidth,height=8cm]{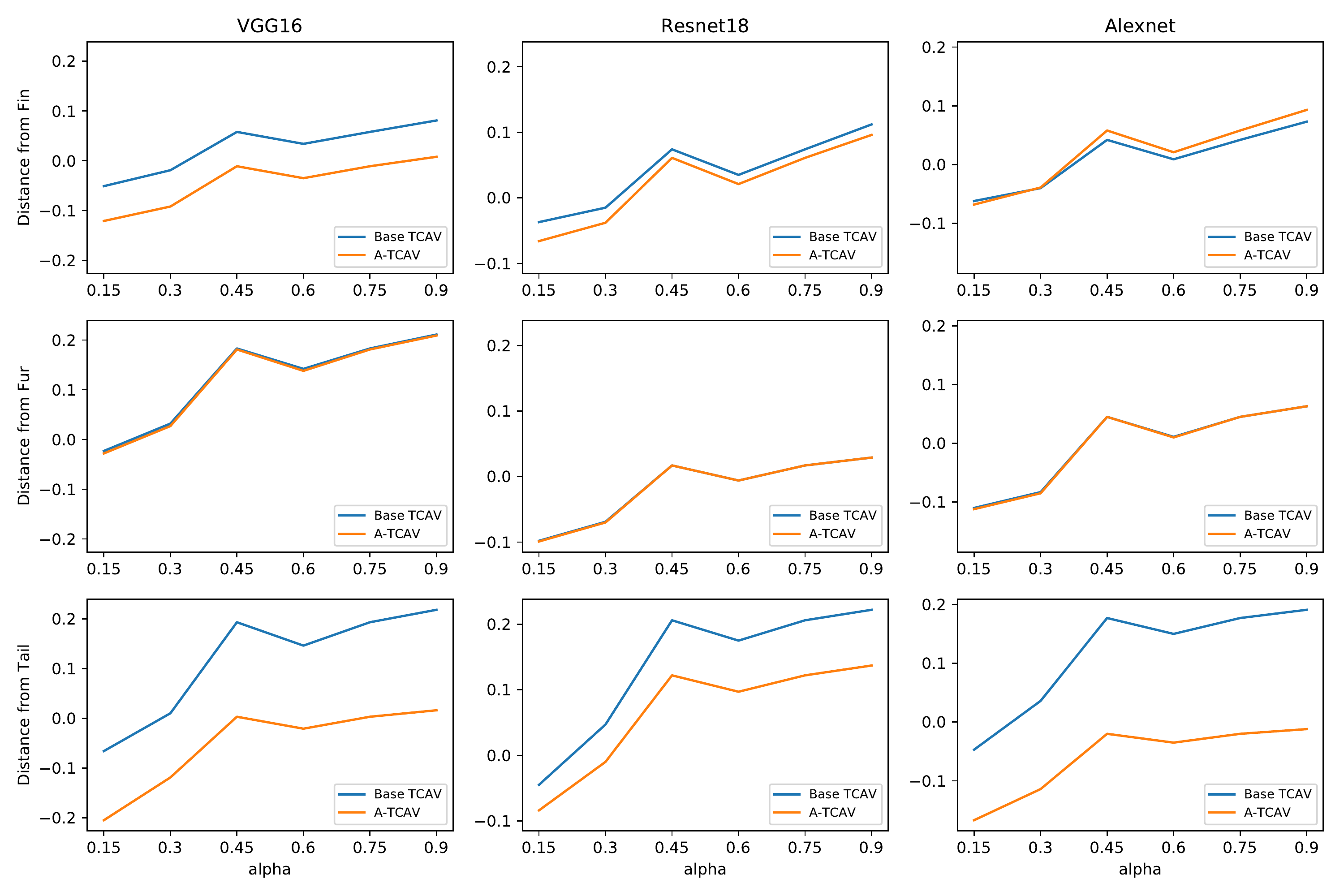} }}%
    \qquad
    \subfloat[seed=21]{{\includegraphics[width=0.46\linewidth,height=8cm]{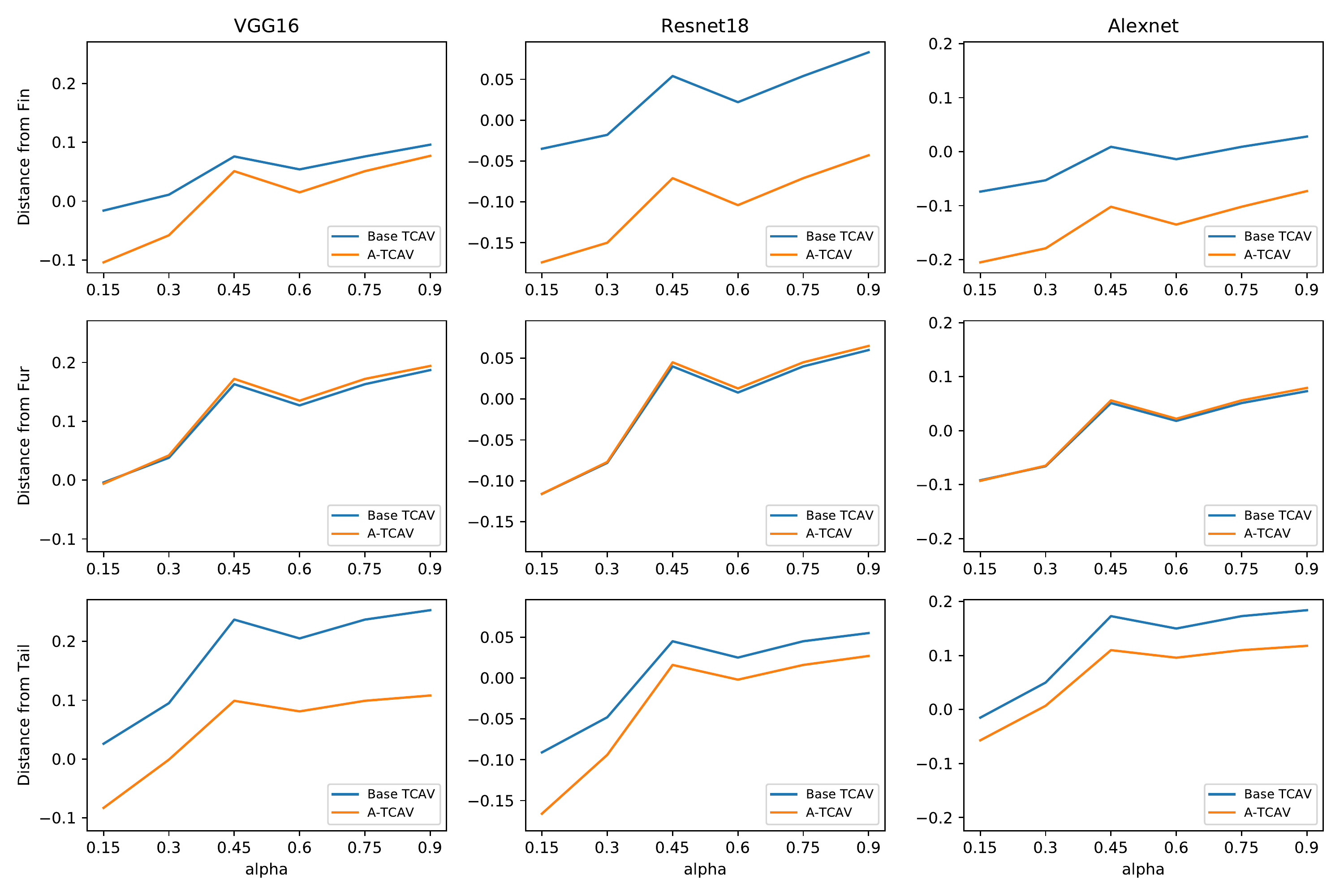} }}%
    \caption{Average distance (defined as inverse of sensitivity) between CAVs and the adversarial dataset with increase level of perturbation in the dataset. Lower numbers mean \textit{closeness} to the CAVs.}%
    \label{fig:sensitivity}
\end{figure}

To investigate empirically, we take the testset images of Imagenet20 dataset and construct adversarial examples using iterative perturbation until a mis-classification has occured \cite{goodfellow2014explaining}. Next, we compare the separation of the adversarial dataset for baseline CAV and A-CAV at the bottleneck layer. As evident from Fig~\ref{fig:sensitivity}, we observe that the proposed method consistently remains \textit{closer} to the concept learned in the intermediate layers --- thus improving the robustness of intermediate layers against adversarial attacks.

\section{Conclusion \& Future Scope}
TCAV is an excellent approach to probe intermediate layers of the neural network and gain human-level understanding of what concept a layer has learned. TCAVs, however, have shortcomings such as effectiveness and robustness that we attempted to overcome.

In the future, we aim to investigate the effects of the following proposals (which are currently outside of the scope of current work): (1) Effect of iterative adversarial updates on the CAV sensitivity, for example the Fast Gradient Sign Attack \cite{goodfellow2014explaining}. Other sophisticated methods could be explored additionally, (2) Consistency measures of Shapley TCAV: \cite{yeh2019concept}, (3) Imposing strictly disjoint sets of positive and negative examples has led to larger standard deviation that A-CAV as we see in Table~\ref{tab:vgg},~\ref{tab:resnet},~\ref{tab:alex}. Methods to relax this criteria and learn a better representation of ``non-concept'' examples would be an interesting topic to explore. 

Currently, CAV helps to learn concepts at layer level which is too broad a definition, since layers contain shared information about concepts from different target categories. Methods to explore and isolate a group of neurons, learning a particular concept in a layer, would be a great future contribution to further probe into the working mechanics of black-box neural networks.

\bibliographystyle{plain}
\bibliography{atcav.bbl}

\begin{thebibliography}{10}

\bibitem{bjorck1994numerics}
{\AA}ke Bj{\"o}rck.
\newblock Numerics of gram-schmidt orthogonalization.
\newblock {\em Linear Algebra and Its Applications}, 197:297--316, 1994.

\bibitem{caruana2015intelligible}
Rich Caruana, Yin Lou, Johannes Gehrke, Paul Koch, Marc Sturm, and Noemie
  Elhadad.
\newblock Intelligible models for healthcare: Predicting pneumonia risk and
  hospital 30-day readmission.
\newblock In {\em Proceedings of the 21th ACM SIGKDD international conference
  on knowledge discovery and data mining}, pages 1721--1730, 2015.

\bibitem{chattopadhay2018grad}
Aditya Chattopadhay, Anirban Sarkar, Prantik Howlader, and Vineeth~N
  Balasubramanian.
\newblock Grad-cam++: Generalized gradient-based visual explanations for deep
  convolutional networks.
\newblock In {\em 2018 IEEE Winter Conference on Applications of Computer
  Vision (WACV)}, pages 839--847. IEEE, 2018.

\bibitem{donoho2004does}
David Donoho and Victoria Stodden.
\newblock When does non-negative matrix factorization give a correct
  decomposition into parts?
\newblock In {\em Advances in neural information processing systems}, pages
  1141--1148, 2004.

\bibitem{farfade2015multi}
Sachin~Sudhakar Farfade, Mohammad~J Saberian, and Li-Jia Li.
\newblock Multi-view face detection using deep convolutional neural networks.
\newblock In {\em Proceedings of the 5th ACM on International Conference on
  Multimedia Retrieval}, pages 643--650, 2015.

\bibitem{gale2018producing}
William Gale, Luke Oakden-Rayner, Gustavo Carneiro, Andrew~P Bradley, and
  Lyle~J Palmer.
\newblock Producing radiologist-quality reports for interpretable artificial
  intelligence.
\newblock {\em arXiv preprint arXiv:1806.00340}, 2018.

\bibitem{geirhos2018imagenet}
Robert Geirhos, Patricia Rubisch, Claudio Michaelis, Matthias Bethge, Felix~A
  Wichmann, and Wieland Brendel.
\newblock Imagenet-trained cnns are biased towards texture; increasing shape
  bias improves accuracy and robustness.
\newblock {\em arXiv preprint arXiv:1811.12231}, 2018.

\bibitem{goodfellow2014explaining}
Ian~J Goodfellow, Jonathon Shlens, and Christian Szegedy.
\newblock Explaining and harnessing adversarial examples.
\newblock {\em arXiv preprint arXiv:1412.6572}, 2014.

\bibitem{kim2016examples}
Been Kim, Rajiv Khanna, and Oluwasanmi~O Koyejo.
\newblock Examples are not enough, learn to criticize! criticism for
  interpretability.
\newblock In {\em Advances in Neural Information Processing Systems}, pages
  2280--2288, 2016.

\bibitem{kim2017interpretability}
Been Kim, Martin Wattenberg, Justin Gilmer, Carrie Cai, James Wexler, Fernanda
  Viegas, and Rory Sayres.
\newblock Interpretability beyond feature attribution: Quantitative testing
  with concept activation vectors (tcav).
\newblock {\em arXiv preprint arXiv:1711.11279}, 2017.

\bibitem{kim2017end}
Jiman Kim and Chanjong Park.
\newblock End-to-end ego lane estimation based on sequential transfer learning
  for self-driving cars.
\newblock In {\em Proceedings of the IEEE Conference on Computer Vision and
  Pattern Recognition Workshops}, pages 30--38, 2017.

\bibitem{kim2017interpretable}
Jinkyu Kim and John Canny.
\newblock Interpretable learning for self-driving cars by visualizing causal
  attention.
\newblock In {\em Proceedings of the IEEE international conference on computer
  vision}, pages 2942--2950, 2017.

\bibitem{kindermans2017learning}
Pieter-Jan Kindermans, Kristof~T Sch{\"u}tt, Maximilian Alber, Klaus-Robert
  M{\"u}ller, Dumitru Erhan, Been Kim, and Sven D{\"a}hne.
\newblock Learning how to explain neural networks: Patternnet and
  patternattribution.
\newblock {\em arXiv preprint arXiv:1705.05598}, 2017.

\bibitem{koh2017understanding}
Pang~Wei Koh and Percy Liang.
\newblock Understanding black-box predictions via influence functions.
\newblock {\em arXiv preprint arXiv:1703.04730}, 2017.

\bibitem{lee1999learning}
Daniel~D Lee and H~Sebastian Seung.
\newblock Learning the parts of objects by non-negative matrix factorization.
\newblock {\em Nature}, 401(6755):788--791, 1999.

\bibitem{lee2018simple}
Kimin Lee, Kibok Lee, Honglak Lee, and Jinwoo Shin.
\newblock A simple unified framework for detecting out-of-distribution samples
  and adversarial attacks.
\newblock In {\em Advances in Neural Information Processing Systems}, pages
  7167--7177, 2018.

\bibitem{liang2017enhancing}
Shiyu Liang, Yixuan Li, and Rayadurgam Srikant.
\newblock Enhancing the reliability of out-of-distribution image detection in
  neural networks.
\newblock {\em arXiv preprint arXiv:1706.02690}, 2017.

\bibitem{lundberg2017unified}
Scott~M Lundberg and Su-In Lee.
\newblock A unified approach to interpreting model predictions.
\newblock In {\em Advances in Neural Information Processing Systems}, pages
  4768--4777, 2017.

\bibitem{ross2017right}
Andrew~Slavin Ross, Michael~C Hughes, and Finale Doshi-Velez.
\newblock Right for the right reasons: Training differentiable models by
  constraining their explanations.
\newblock {\em arXiv preprint arXiv:1703.03717}, 2017.

\bibitem{selvaraju2017grad}
Ramprasaath~R Selvaraju, Michael Cogswell, Abhishek Das, Ramakrishna Vedantam,
  Devi Parikh, and Dhruv Batra.
\newblock Grad-cam: Visual explanations from deep networks via gradient-based
  localization.
\newblock In {\em Proceedings of the IEEE international conference on computer
  vision}, pages 618--626, 2017.

\bibitem{smilkov2017smoothgrad}
Daniel Smilkov, Nikhil Thorat, Been Kim, Fernanda Vi{\'e}gas, and Martin
  Wattenberg.
\newblock Smoothgrad: removing noise by adding noise.
\newblock {\em arXiv preprint arXiv:1706.03825}, 2017.

\bibitem{springenberg2014striving}
Jost~Tobias Springenberg, Alexey Dosovitskiy, Thomas Brox, and Martin
  Riedmiller.
\newblock Striving for simplicity: The all convolutional net.
\newblock {\em arXiv preprint arXiv:1412.6806}, 2014.

\bibitem{sundararajan2017axiomatic}
Mukund Sundararajan, Ankur Taly, and Qiqi Yan.
\newblock Axiomatic attribution for deep networks.
\newblock In {\em Proceedings of the 34th International Conference on Machine
  Learning-Volume 70}, pages 3319--3328. JMLR. org, 2017.

\bibitem{ustun2013supersparse}
Berk Ustun, Stefano Traca, and Cynthia Rudin.
\newblock Supersparse linear integer models for interpretable classification.
\newblock {\em arXiv preprint arXiv:1306.6677}, 2013.

\bibitem{yeh2019concept}
Chih-Kuan Yeh, Been Kim, Sercan~O Arik, Chun-Liang Li, Pradeep Ravikumar, and
  Tomas Pfister.
\newblock On concept-based explanations in deep neural networks.
\newblock {\em arXiv preprint arXiv:1910.07969}, 2019.

\bibitem{zeiler2014visualizing}
Matthew~D Zeiler and Rob Fergus.
\newblock Visualizing and understanding convolutional networks.
\newblock In {\em European conference on computer vision}, pages 818--833.
  Springer, 2014.

\bibitem{zhang2018interpretable}
Kai Zhang, Xiyang Liu, Fan Liu, Lin He, Lei Zhang, Yahan Yang, Wangting Li,
  Shuai Wang, Lin Liu, Zhenzhen Liu, et~al.
\newblock An interpretable and expandable deep learning diagnostic system for
  multiple ocular diseases: qualitative study.
\newblock {\em Journal of medical Internet research}, 20(11):e11144, 2018.

\end{thebibliography}
\end{document}